\documentclass[10pt,twocolumn,letterpaper]{article}

\usepackage{cvpr} 
\usepackage[numbers]{natbib}
\definecolor{cvprblue}{rgb}{0.21,0.49,0.74}
\usepackage[pagebackref,breaklinks,colorlinks,allcolors=cvprblue]{hyperref}

\usepackage{booktabs}
\def\paperID{51} 
\def\confName{CV4Animals Workshop Track 1}
\def\confYear{2025}

\title{Tracking the Flight: Exploring a Computational Framework for Analyzing Escape Responses in Plains Zebra (\textit{Equus quagga})}

\author{
Isla Duporge$^{1*}$ \quad
Sofía Miñano$^{2*}$ \quad
Nikoloz Sirmpilatze$^{2}$ \quad
Igor Tatarnikov$^{2}$ \\
Scott Wolf$^{1}$ \quad
Adam L. Tyson$^{2,3}$ \quad
Daniel Rubenstein$^{1}$ \\
$^1$Princeton University \quad
$^2$Sainsbury Wellcome Centre, University College London \\
$^3$Gatsby Computational Neuroscience Unit, University College London \\
$^*$These authors contributed equally \\
\texttt{Isla.duporge@princeton.edu, s.minano@ucl.ac.uk}
}
\begin{document}
\maketitle
\begin{abstract}

Ethological research is increasingly aided by the growing affordability and accessibility of drones, which enable the capture of high-resolution footage of animal movement at fine spatial and temporal scales. However, analyzing such footage presents the technical challenge of separating animal movement from drone motion. While non-trivial, computer vision techniques, such as image registration and Structure-from-Motion (SfM), offer practical solutions. For conservationists, open-source tools that are user-friendly, require minimal setup, and deliver timely results are especially useful for efficient data interpretation. This study evaluates three approaches: a bioimaging-based registration technique, an SfM pipeline, and a hybrid interpolation method. We apply these to a recorded escape event involving 44 plains zebras (\textit{Equus quagga}), captured in a single drone video. Using the best-performing method, we extract individual trajectories and identify key behavioral patterns: increased alignment (polarization) during escape, a brief widening of spacing just before stopping, and tighter coordination near the group’s center. These insights highlight the method’s effectiveness and the potential to scale to larger datasets, contributing to broader investigations of collective animal behavior.

\end{abstract}    
\section{Introduction}
\label{sec:intro}
Among the evolutionary drivers of social behavior, marginal predation is widely regarded as a key force shaping both grouping tendencies and centripetal instincts—the tendency of individuals to move toward the center of a group \cite{Vine1971Flocking}. This evolutionary pressure can be seen in the movement patterns of herding species such as wildebeest, where group shape varies with speed: wave-like formations occur during walking, while faster movement produces more linear, columnar arrangements \cite{GUERON199685}. These behavioral patterns align with the Selfish Herd hypothesis, which suggests that individuals move toward the center of a group when threatened, thereby reducing their personal risk of predation \cite{Hamilton1971-pp, Viscido2002-wh}. This inward movement has been observed across a wide range of species \cite{Heard1992Wolf, Spieler1999Aggregation, Watt1997Toad, Ens1993Flocking} and is clearly illustrated in studies of sheepdog–sheep interactions \cite{Jadhav2024Sheep}.

Group living provides  benefits such as energetic efficiency and improved predator detection. These advantages include collective vigilance, as described by the “many eyes” hypothesis, and predator confusion, which may be especially effective in striped species like zebras \cite{Viscido2002-wh}. Plains zebra (\textit{Equus quagga}), the focus of this study, typically form harems of 5–20 individuals, which can temporarily aggregate into larger herds of over 50. These highly social animals maintain cohesion and defend against predators such as lions and hyenas through vocalizations, grooming, and coordinated responses to threats \cite{CHEN2021100001, Rubenstein2023}. However, the specific strategies zebras use to coordinate their escape behavior in response to predators remain largely unexamined.

\subsection{Collective escape behavior}

Collective escape dynamics have been widely studied in gregarious species, particularly in bird flocks \cite{Portugal2020-be, Papadopoulou2022-ys} and fish groups, including polarized schools and less-coordinated shoals \cite{Miller2012Zebrafish, Landeau1986-iy}. These species navigate in three dimensions—birds adjust altitude and fish change depth. Terrestrial mammals, restricted to two dimensions, exhibit escape dynamics that are fundamentally different and less well understood.

It remains unclear how aware individuals are of their relative position in a group or whether they prioritize personal advantage over collective outcomes. At high densities, flocks and schools can become more coordinated and less responsive to perturbation under predation \cite{Cavagna2010StarlingFlocks, Handegard2012GroupHunting, Herbert-Read2015-cn}. Tracking neighbors during escape may impair alignment due to cognitive demands, but this likely varies by group position \cite{Rubenstein2023}—followers must attend more to neighbors, while leaders may focus on navigating escape routes \cite{HERBERTREAD2015R1127}.

Most studies of escape behavior focus on single species in lab settings or rely on theoretical models lacking field validation, limiting cross-species generalization. Escape dynamics in terrestrial mammals are less studied, partly due to difficulties in tracking fast-moving groups and observing rare, unpredictable predator events \cite{Papadopoulou2022-qu}.

However, emerging technologies—such as drones \cite{duporge2024}, satellites \cite{Wu2023}, and advances in automated tracking and pose estimation \cite{Pereira2022SLEAP}—now allow detailed observation of group behavior in the wild. The growing accessibility of machine learning tools and affordable hardware presents new opportunities to investigate these behaviors across diverse species in natural settings.


\section{Methods}
\label{sec:methods}
\subsection{Data collection}
Drone videos were collected during controlled behavioral experiments conducted in Mpala (Kenya), in an environment of savannah and dry woodland habitat. The area is unfenced, allowing wildlife to roam freely. All experiments discussed here were conducted with plains zebra (Equus quagga) in their natural, unconstrained environment.

Video data were recorded using two quadcopter drone models: the DJI Mavic 3 Pro, capturing footage in 4K Ultra HD (3840 × 2160 pixels) and the DJI Mini 2 SE, recording in Full HD (1920 × 1080). All data was recorded at 29.97 fps. The video used as case example in this study was captured with the DJI Mavic 3 Pro. Prior to data collection, test flights were conducted to assess potential disturbance to the zebras, following established guidelines for minimizing drone-related behavioral interference \cite{Dronedisturbance}. Zebras exhibited no signs of disturbance at altitudes above 60 m, and all experimental flights were conducted at or above this threshold, with the majority occurring at 100 m altitude.

Experimental trials were ran to simulate a predation event. Opportunistically encountered zebra herds were approached by two to three researchers walking briskly in a triangular formation, mimicking the coordinated approach of a hunting lion coalition. This induced a flight response in the herd, which was recorded from the drone until the group resumed a relaxed, loosely aggregated state and began grazing again. The experiment was repeated across different zebra herds over a five-day period (January 13–22, 2025), resulting in a dataset of 41 videos. Herd sizes varied between 3 and 44 individuals across trials. No same group was approached more than once a day, but it is not known how many different herds are included in the dataset as individual identification was not possible.
\subsection{Pose estimation and tracking}
All videos were manually reviewed, 41 of those recorded were selected as suitable for inclusion. Out-of-focus videos or those with frequent changes in flight altitude and camera angle were excluded. The videos were shared with a professional labeling company (\url{https://labelyourdata.com/}) which provided ground-truth data to train a pose estimation using the software SLEAP \cite{Pereira2022SLEAP}. Keypoints were defined on the head and tail of each zebra, and 1963 frames (approximately 1/5 of the video) were manually annotated. We trained a multi-animal top-down SLEAP model, consisting of a centroid and a centered-instance model, with a U-Net backbone. We used the flow tracker for cross-frame identity. The validation set consisted of 10\% of the annotated frames (196 frames), which we used to evaluate both the centroid model (mean localization error $d=0.37$ pixels), and the centered-instance model (mean localization error $d=1.33$ pixels). Note that as is common in the scientific use of pose estimation packages such as SLEAP and due to the relatively limited availability of labeled data, our goal is to produce a model that performs well in our dataset of interest, rather than a model that generalizes well in a wide range of unseen scenarios.

For the analysis presented here, we selected a 3.5-minute video from the dataset that captures a group of 44 individuals during four escape waves. We treat the entire video as a single escape event, initiated by a single provocation from the researchers at the beginning. During this event, the herd exhibits four distinct escape waves, each consisting of a collective run followed by a stop.

\subsection{Unwrapping zebra tracks}
We considered three different approaches to "unwrap" the trajectories of the moving zebras in the selected video, using open-source freely available software and off-the-shelf models. We use "unwrapping" to refer to the process of transforming the trajectories of the animals from a moving image coordinate system fixed to the drone, to a world coordinate system that is fixed to the ground. 

As a baseline, the first approach uses an image registration algorithm to compute frame-to-frame transformations. The second and third approaches rely on Structure-from-Motion (SfM) to estimate camera poses at selected keyframes, which we then interpolated using two different methods. The SfM-based methods are inspired by a related work \cite{koger2023} that used commercial SfM software to generate anchor frames.

\subsubsection{Image registration using `itk-elastix`}
We repurposed an existing image registration algorithm - originally developed for medical imaging - to compute the transform from each frame to the previous frame. 

Specifically, we used `itk-elastix', a toolbox for rigid and nonrigid registration of images \cite{klein2009elastix,Ntatsis2023-ik,shamonin2014elastix}, that is itself based on the Insight Segmentation and Registration Toolkit (ITK) \cite{ibanez2025itk_14867643}. We used its transformation module to compute the 2D rigid transformations $[R | \vec{t}]_{f, f-1}$ from the image in frame $f$ to the image in frame $f-1$ that minimize the mutual information metric \cite{thevenaz2000optimization}. We masked out the pixels around the individual zebras from the registration process. We then computed the transforms from the image coordinate system of frame $f$ to the one of the initial frame $f=0$, by composing the frame-to-frame transformations:
\[
[R | \vec{t}]_{f, f=0} = \prod_{j=1}^{f} [R | \vec{t}]_{j, j-1}
\]

In this case, we defined the world coordinate system as the coordinate system that is parallel to the image coordinate system at frame $f=0$ and is fixed to the ground. By applying the computed transforms to the trajectory data $X$ in the image coordinate system ($ICS$) of each frame $f$, we obtained the corresponding trajectories in the world coordinate system ($WCS$):
\[
X_{WCS} = Q^T  [R | \vec{t}]_{f, f=0} Q X_{ICS, f}
\]

Note that this approach based on frame-to-frame transformations is prone to cumulative error, especially over longer or complex flight paths. However, it may still be useful in scenarios where the imagery remains relatively stable, such as when the drone is hovering in place. In this case study, we used this method as a baseline for comparison.

\begin{figure*}[!ht]
  \centering
    \includegraphics[width=2\columnwidth]{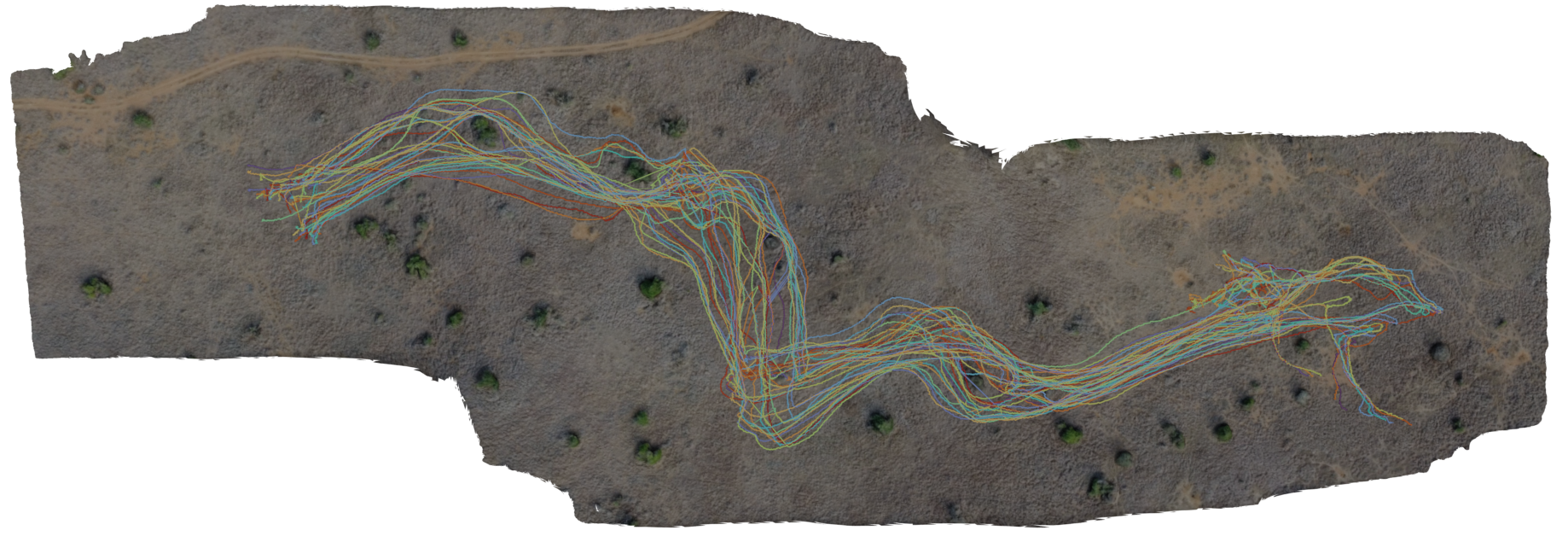}
    \caption{Stitched drone flight path overlaid with the movement trajectories of 44 individual zebra within the observed herd. The trajectories are computed using the SfM linearly-interpolated approach.}
    \label{fig:stitched-image-with-tracks}
\end{figure*}

\subsubsection{Interpolated Structure-from-Motion}
We used tools from the OpenDroneMap ecosystem \cite{authors2020odm} to estimate the camera poses at specified keyframes. Specifically, we applied the OpenSfM package to perform Structure-from-Motion on every 20th frame of the video, plus the first and the last frames. To minimize the impact of moving zebras on the reconstruction process, we masked out their pixels from the input to the SfM pipeline. The 316 images were processed in 47 min 53 s using a consumer-level GPU (NVIDIA GeForce RTX 4080 16GB), and we obtained an average reprojection error of 0.58 pixels.

We then interpolated the camera poses at the missing frames. For the camera translation, we linearly interpolated the position assuming constant velocity motion. For the camera orientation, we considered two methods:
\begin{itemize}
    \item Spherical Linear Interpolation (SLERP) \cite{shoemake1985animating}, which assumes constant angular velocity between keyframes;
    \item Image registration-based rotation, which estimates the rotation of the image plane by registering each missing frame to its nearest preceding keyframe.
\end{itemize}
Note that the SLERP interpolation operates purely on keyframe poses, while the registration-based method makes use of pixel content. In both cases, once all camera poses are defiend, we estimated the 3D trajectories by computing the intersection between the viewing ray — originating from the camera origin and passing through the relevant pixel — and the best-fitting plane to the OpenDroneMap 2.5D mesh. For the behavioural analysis, we used the 2D trajectories resulting from the projection onto the best-fitting plane.

\subsubsection{Comparison of approaches}
To evaluate the performance of the three proposed approaches, we detected and tracked 45 trees in the selected video. Trees serve as effective reference points because they are visually distinctive and should be stationary in the world coordinate system. To extract their trajectories in the image coordinate system, we used `deepforest` \cite{weinstein2020deepforest, weinstein2019individual} for detection and the BotSort algorithm implementation from boxmot \cite{mikel_brostrom_2023_8132989} for tracking. We filtered the resulting trajectories for quality, retaining only those with more than 400 data points and excluding any with unrealistic frame-to-frame jumps (greater than 10 pixels per frame).

The resulting trajectories were then unwrapped using each of the methods described above. Since trees are static, any dispersion in their unwrapped trajectories in the world coordinate system serves as a proxy for error. To quantify this dispersion, we computed the mean distance of each trajectory to its centroid in world coordinates, weighted by the number of samples. To use comparable units across methods, we normalized the distance values by the median zebra body length (computed from the zebra trajectories processed with the same method and cleaned as described in Section \ref{ssec:herd_behavior_metrics}). The computed metrics are shown in Table \ref{tab:error_metrics}, and the metrics per tree and method can be found in the supplementary material (tables \ref{tab:si-itk-all-tree-distances} to \ref{tab:si-sfm-itk-tree-distances}).

\begin{table}[ht]
\centering
\caption{Error metrics for each of the unwrapping methods. The error is computed using the unwrapped tree trajectories, as the weighted average distance of each tree to its centroid, normalised by the median zebra body-length.}
\label{tab:error_metrics}
\centering
\footnotesize
\begin{tabular}{@{}lccc@{}}
\toprule
\textbf{Method} & \textbf{Weighted average distance} \\
\midrule
Image Registration                    & 0.910  \\
\textbf{SfM + Linear Interpolation}            & \textbf{0.275}  \\
SfM + Registration-Based Rotation     & 0.299 \\
\bottomrule
\end{tabular}
\end{table}

As expected, in this data the frame-to-frame image registration approach yields a mean weighted error close to a zebra’s body-length, making it less suitable for fine-scale analysis. In contrast, the SfM-based methods perform better, with the linearly interpolated variant achieving the lowest error. Based on these results, we use the SfM with linear interpolation for the subsequent behavioural analysis.

\subsection{Herd behavior metrics}
\label{ssec:herd_behavior_metrics}
For the behavioral analysis, we first processed the unwrapped keypoint tracks of the zebras using the open-source Python package `movement' \cite{sirmpilatze2025_15026249}. Keypoints with confidence scores below 0.9 (as reported by SLEAP) were discarded. We estimated body length as the median distance between the head and tail keypoints across the full clip, and excluded any points that moved more than two body lengths between consecutive frames. 
We then computed herd behavior metrics also using the functionalities of the `movement' package \cite{sirmpilatze2025_15026249}. 

Body orientation was inspected via the individuals' body vectors, defined from tail to head keypoints. We removed outlier body vectors from the analysis — those deviating more than $\pm 2$ standard deviations from the mean length. We computed the alignment of each individual's body vector with the mean body orientation, as well as the polarization metric of the full herd at each timestep. The polarization was computed at each timestep as the norm of the mean unit body vector across individuals. Polarization ranges from 0 (fully dispersed) to 1 (fully aligned).

We also computed each zebra's centroid (midpoint between head and tail) and derived centroid speed in units of body lengths per second. Additionally, we measured the distance between all unique pairs of centroids, and derived the mean and the maximum inter-individual distance at each timestep. The position in the herd is measured as distance from the herd's centroid.

\begin{figure}
  \centering
    \includegraphics[width=\columnwidth]{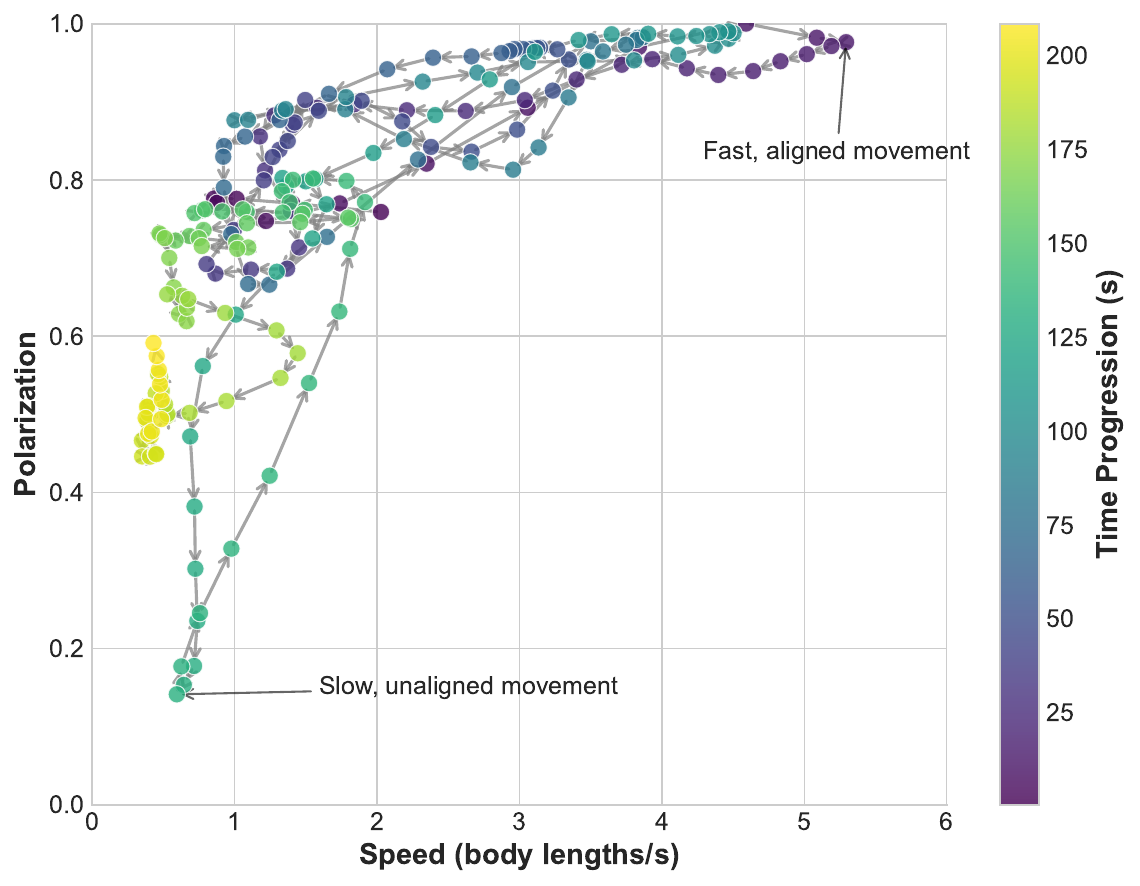}
    \caption{Scatter plot showing co-evolution of average group speed (body lengths/s) and alignment (polarization) in in $\sim$1-second intervals (30 frames), colored by time. Arrows trace temporal progression, highlighting shifts between fast, aligned and slow, unaligned movement. Smoothed with a Savitzky-Golay filter (window size 7); key regions annotated.}
    \label{fig:polarization-speed}
\end{figure}
The supporting code for the complete analysis is available at \href{https://github.com/neuroinformatics-unit/zebras-stitching}{https://github.com/neuroinformatics-unit/zebras-stitching}. The repository is not a fully-fledged software package, but rather as a collection of prototyping notebooks to explore the three different approaches considered here and their trade-offs. We hope these notebooks can serve as a starting point for researchers to get familiar with the problem and the existing tools, or for further development of a robust pipeline based on open-source freely available software.

\section{Results}
\label{sec:Results}

\begin{figure}[!ht]
  \centering
    \includegraphics[width=\columnwidth]{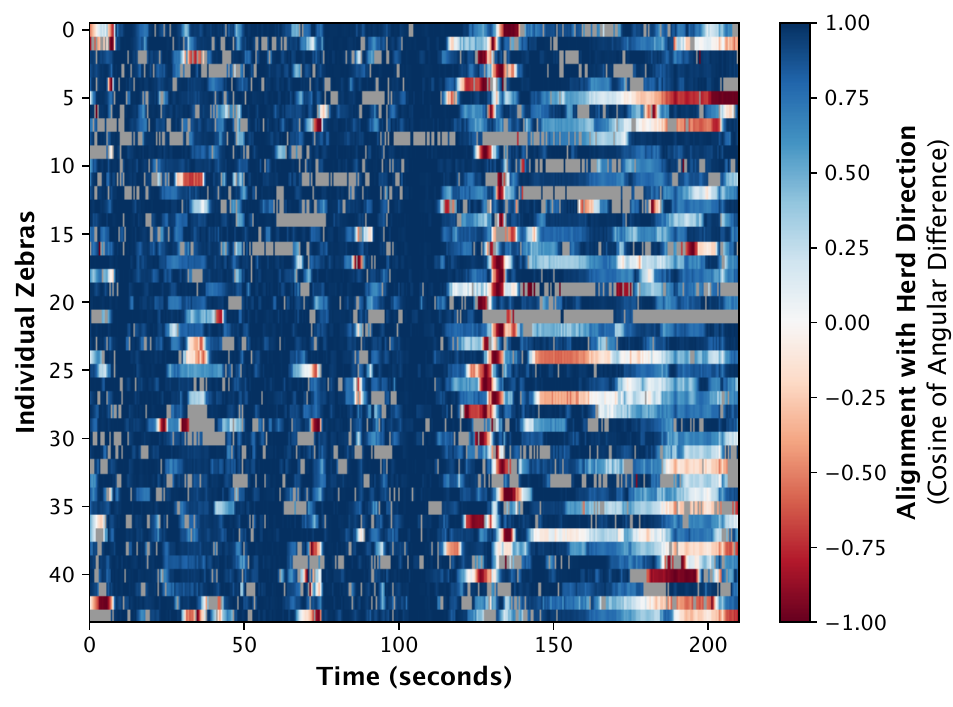}
    \caption{Plot showing individual zebra alignment with the herd’s mean direction over time. Each row is a zebra; color indicates the cosine of the angular difference between its orientation and the group’s (red: opposite, white: perpendicular, blue: aligned, grey: missing).Values range from -1 (opposite) to 1 (aligned).}
    \label{fig:zebra_individual_alignment}
\end{figure}

\begin{figure}[!ht]
  \centering
    \includegraphics[width=\columnwidth]{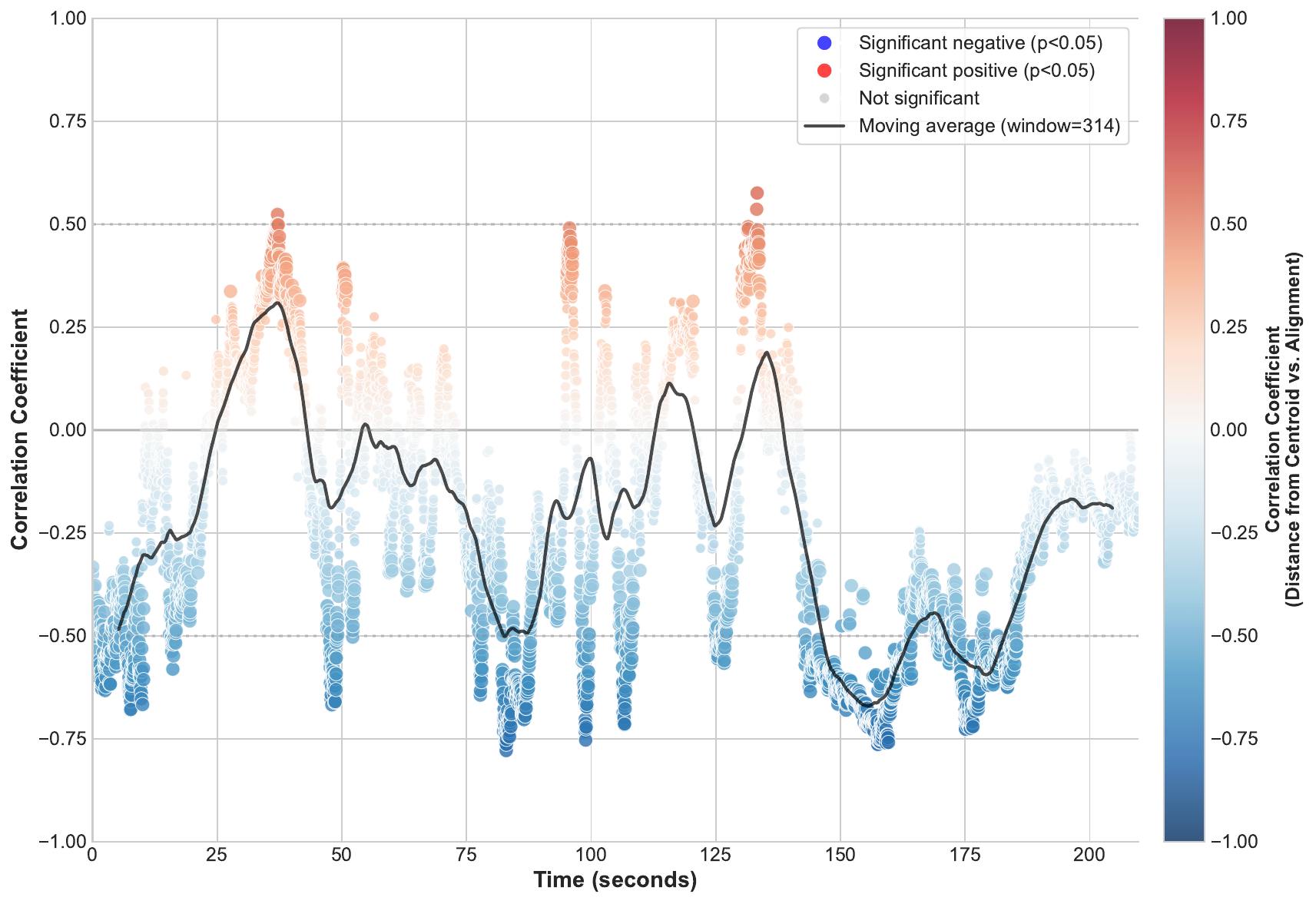}
    \caption{Frame-by-frame Pearson correlation between each zebra’s distance from the herd centroid and its alignment with the average herd direction. Each point shows whether individuals farther from the center are more or less aligned in that frame.}
    \label{fig:zebra_frame_level_position_alignment_correlation}
\end{figure}

Although the analysis is based on a single video and therefore not sufficient to generalize about plains zebra escape behavior broadly, it serves as a proof of concept demonstrating the types of movement metrics that can be extracted using the selected method. In this exemplar case, we observe a high degree of alignment (polarization) among individuals, which decreases during pauses in movement. Prior to stopping, individuals exhibit an increase in inter-individual distance, potentially reflecting enhanced vigilance behavior. Despite distinct waves of movement initiation, nearest-neighbor distances remain relatively stable throughout the event (Fig.~\ref{fig:stitched-image-with-tracks}). Alignment increases with individual speed (Fig.~\ref{fig:polarization-speed}) and the herd maintains cohesion, with stable lateral positioning observed approximately two-thirds into the escape sequence (Fig.~\ref{fig:zebra_individual_alignment}). Zebras positioned near the center of the group show consistently higher alignment than those on the periphery (Fig.~\ref{fig:zebra_frame_level_position_alignment_correlation}). These findings demonstrate the potential of this method to characterize escape dynamics in animal groups and provide a scalable framework for future comparative analyses across events and taxa. 



\section{Discussion}
\label{sec:Discussion}
Our behavioural analysis of this escape run, integrating video observations with quantitative metrics, reveals that zebras dynamically balance individual decision-making with social cues to maintain group cohesion under perceived threat. Patterns of alignment, spacing, and group structure suggest a flexible coordination strategy, with different individuals leading at various moments. A transient increase in spacing before stopping may enhance vigilance or reduce collision risk. Contrary to classic ‘selfish herd’ models \cite{Hamilton1971-pp}, we do not observe individuals consistently moving toward the centre. While strong group alignment aligns with prior findings \cite{Rubenstein2023}, our results further reveal greater alignment among centrally positioned individuals than those on the periphery.

These findings are based on a single escape event used to illustrate our method, which can be extended to the full dataset of 44 videos to assess the consistency of escape dynamics across group sizes and contexts. Although releasing predators is not feasible, a coordinated approach by three researchers—mimicking a lion’s hunting strategy—offers an ecologically realistic simulation. Robotic predators offer a promising alternative \cite{Rubenstein2023, Bonnet2019Robots}.

We tested three methods to “unwrap” animal trajectories from drone footage, prioritising user-friendly tools with minimal parameter tuning. Using trees as ground references, we found SfM-based methods outperformed frame-to-frame image registration. The linearly interpolated SfM approach yielded the best accuracy, with an average dispersion of 0.275 body lengths, while the image registration method performed less well, though it may suit scenarios with minimal drone motion.

In future work, incorporating geotagged data could further improve SfM reconstructions. Our framework can also benchmark emerging methods in conservation contexts. A promising model is VGGT \cite{wang2025vggt}, which rapidly estimates 3D scene attributes from single or multiple views. A robust validation framework will be key to scaling and comparing such methods for behavioural analysis.

\subsection{Ethical Considerations}
The data was gathered in accordance with Research License No. NACOSTI/P/22/18214 and the guidelines set forth by the Institutional Animal Care and Use Committee under Princeton University permission Nos. IACUC 1842F and 1835F. 

\subsection{Acknowledgments}
We are grateful to Craig Marshall, Erin Flowers, and all students of the 'Techniques in Field Biology Course' for their valuable assistance with data collection. The NSF OAC 2118240 Imageomics Institute award supported D.R. and I.D.

{
    \small
    \bibliographystyle{ieeenat_fullname}
    \bibliographystyle{unsrt}

    \bibliography{main}
}
\clearpage
\setcounter{page}{1}
\maketitlesupplementary

\begin{table}
\caption{Distance to centroid normalized by median zebra body length. Approach: image registration frame-to-frame. Weighted normalized mean: 0.910 (body length units). "std" stands for standard deviation.}
\label{tab:si-itk-all-tree-distances}
\begin{tabular}{lrrrrr}
\toprule
 tree ID & mean & max & min & std & samples \\
\midrule
1 & 0.319 & 3.150 & 0.021 & 0.427 & 413 \\
2 & 0.838 & 6.103 & 0.028 & 0.700 & 1799 \\
3 & 0.821 & 3.018 & 0.304 & 0.304 & 1600 \\
5 & 0.311 & 0.991 & 0.010 & 0.203 & 414 \\
6 & 0.410 & 2.437 & 0.033 & 0.289 & 483 \\
30 & 1.379 & 4.428 & 0.047 & 0.600 & 2055 \\
31 & 0.880 & 3.361 & 0.094 & 0.444 & 1180 \\
37 & 0.604 & 1.572 & 0.054 & 0.268 & 1082 \\
40 & 0.699 & 1.790 & 0.021 & 0.351 & 449 \\
47 & 0.661 & 4.148 & 0.101 & 0.477 & 1501 \\
50 & 0.511 & 1.019 & 0.018 & 0.186 & 607 \\
68 & 0.998 & 5.188 & 0.013 & 0.711 & 2044 \\
107 & 0.263 & 1.077 & 0.014 & 0.184 & 593 \\
124 & 1.206 & 3.210 & 0.133 & 0.634 & 865 \\
134 & 0.768 & 3.873 & 0.157 & 0.556 & 535 \\
151 & 1.213 & 3.140 & 0.077 & 0.505 & 898 \\
176 & 1.088 & 2.793 & 0.057 & 0.604 & 678 \\
185 & 0.884 & 2.976 & 0.050 & 0.584 & 799 \\
189 & 1.743 & 4.257 & 0.157 & 0.931 & 1467 \\
192 & 0.802 & 2.414 & 0.139 & 0.472 & 678 \\
194 & 0.847 & 3.191 & 0.150 & 0.614 & 747 \\
215 & 1.214 & 3.033 & 0.122 & 0.691 & 1056 \\
218 & 1.645 & 3.778 & 0.008 & 0.829 & 1223 \\
224 & 1.136 & 3.025 & 0.122 & 0.614 & 748 \\
230 & 1.537 & 3.265 & 0.140 & 0.737 & 866 \\
262 & 1.887 & 5.031 & 0.461 & 1.101 & 1365 \\
307 & 0.739 & 1.712 & 0.030 & 0.501 & 547 \\
308 & 0.641 & 1.790 & 0.062 & 0.342 & 516 \\
317 & 0.634 & 1.699 & 0.153 & 0.300 & 420 \\
318 & 0.973 & 2.974 & 0.026 & 0.426 & 551 \\
336 & 0.963 & 2.651 & 0.004 & 0.609 & 2759 \\
337 & 0.484 & 1.744 & 0.032 & 0.400 & 526 \\
344 & 0.505 & 1.878 & 0.097 & 0.439 & 530 \\
366 & 0.898 & 2.028 & 0.149 & 0.347 & 1671 \\
371 & 0.755 & 1.297 & 0.020 & 0.272 & 631 \\
376 & 0.711 & 1.531 & 0.022 & 0.331 & 808 \\
384 & 0.723 & 2.922 & 0.091 & 0.372 & 1313 \\
465 & 0.256 & 0.580 & 0.014 & 0.121 & 488 \\
469 & 0.567 & 1.398 & 0.019 & 0.318 & 903 \\
473 & 0.219 & 0.559 & 0.040 & 0.113 & 401 \\
476 & 0.566 & 1.354 & 0.012 & 0.341 & 922 \\
479 & 0.516 & 1.207 & 0.014 & 0.294 & 901 \\
487 & 0.443 & 0.957 & 0.061 & 0.217 & 722 \\
496 & 0.437 & 1.121 & 0.008 & 0.272 & 631 \\
498 & 0.377 & 0.776 & 0.096 & 0.187 & 494 \\
\bottomrule
\end{tabular}
\end{table}

\begin{table}
\caption{Distance to centroid normalized by median zebra body length. Approach: SfM linearly interpolated. Weighted normalized mean: 0.275 (body length units). "std" stands for standard deviation.}
\label{tab:si-sfm-interp-tree-distances}
\begin{tabular}{lrrrrr}
\toprule
 tree ID & mean & max & min & std & samples \\
\midrule
1 & 0.251 & 2.783 & 0.001 & 0.349 & 413 \\
2 & 0.385 & 3.580 & 0.013 & 0.355 & 1799 \\
3 & 0.382 & 2.660 & 0.010 & 0.204 & 1600 \\
5 & 0.165 & 0.502 & 0.007 & 0.081 & 414 \\
6 & 0.202 & 2.283 & 0.000 & 0.237 & 483 \\
30 & 0.365 & 1.390 & 0.002 & 0.175 & 2055 \\
31 & 0.238 & 1.538 & 0.008 & 0.154 & 1180 \\
37 & 0.245 & 0.951 & 0.001 & 0.141 & 1082 \\
40 & 0.269 & 0.927 & 0.015 & 0.175 & 449 \\
47 & 0.176 & 2.039 & 0.010 & 0.176 & 1501 \\
50 & 0.178 & 0.458 & 0.013 & 0.082 & 607 \\
68 & 0.416 & 1.851 & 0.010 & 0.205 & 2044 \\
107 & 0.213 & 0.784 & 0.009 & 0.131 & 593 \\
124 & 0.235 & 0.775 & 0.006 & 0.118 & 865 \\
134 & 0.280 & 1.458 & 0.022 & 0.157 & 535 \\
151 & 0.251 & 0.791 & 0.011 & 0.138 & 898 \\
176 & 0.274 & 2.372 & 0.014 & 0.241 & 678 \\
185 & 0.248 & 0.869 & 0.016 & 0.170 & 799 \\
189 & 0.562 & 1.710 & 0.023 & 0.317 & 1467 \\
192 & 0.194 & 0.895 & 0.007 & 0.143 & 678 \\
194 & 0.263 & 1.743 & 0.004 & 0.226 & 747 \\
215 & 0.385 & 1.584 & 0.023 & 0.291 & 1056 \\
218 & 0.590 & 1.825 & 0.029 & 0.321 & 1223 \\
224 & 0.196 & 0.537 & 0.009 & 0.097 & 748 \\
230 & 0.374 & 1.354 & 0.019 & 0.281 & 866 \\
262 & 0.607 & 1.337 & 0.006 & 0.272 & 1365 \\
307 & 0.213 & 0.549 & 0.003 & 0.108 & 547 \\
308 & 0.140 & 0.410 & 0.012 & 0.068 & 516 \\
317 & 0.177 & 0.543 & 0.007 & 0.089 & 420 \\
318 & 0.322 & 1.341 & 0.011 & 0.217 & 551 \\
336 & 0.185 & 0.521 & 0.002 & 0.085 & 2759 \\
337 & 0.156 & 0.650 & 0.007 & 0.094 & 526 \\
344 & 0.100 & 0.290 & 0.017 & 0.042 & 530 \\
366 & 0.273 & 0.807 & 0.008 & 0.154 & 1671 \\
371 & 0.125 & 0.558 & 0.004 & 0.084 & 631 \\
376 & 0.174 & 0.602 & 0.012 & 0.098 & 808 \\
384 & 0.203 & 2.057 & 0.013 & 0.170 & 1313 \\
465 & 0.114 & 0.390 & 0.008 & 0.043 & 488 \\
469 & 0.074 & 0.276 & 0.005 & 0.048 & 903 \\
473 & 0.091 & 0.210 & 0.010 & 0.048 & 401 \\
476 & 0.101 & 0.426 & 0.004 & 0.054 & 922 \\
479 & 0.060 & 0.203 & 0.002 & 0.033 & 901 \\
487 & 0.250 & 0.912 & 0.010 & 0.139 & 722 \\
496 & 0.116 & 0.307 & 0.003 & 0.061 & 631 \\
498 & 0.106 & 0.275 & 0.009 & 0.049 & 494 \\
\bottomrule
\end{tabular}
\end{table}

\begin{table}
\caption{Distance to centroid normalized by median zebra body length. Approach: SfM and image registration. Weighted normalized mean: 0.299 (body length units). "std" stands for standard deviation.}
\label{tab:si-sfm-itk-tree-distances}
\begin{tabular}{lrrrrr}
\toprule
 tree ID & mean & max & min & std & samples \\
\midrule
1 & 0.294 & 2.950 & 0.016 & 0.358 & 413 \\
2 & 0.430 & 3.583 & 0.014 & 0.361 & 1799 \\
3 & 0.392 & 2.554 & 0.005 & 0.225 & 1600 \\
5 & 0.187 & 0.614 & 0.002 & 0.107 & 414 \\
6 & 0.218 & 2.294 & 0.009 & 0.244 & 483 \\
30 & 0.394 & 1.422 & 0.006 & 0.196 & 2055 \\
31 & 0.261 & 1.600 & 0.006 & 0.190 & 1180 \\
37 & 0.267 & 1.057 & 0.003 & 0.171 & 1082 \\
40 & 0.305 & 0.931 & 0.019 & 0.179 & 449 \\
47 & 0.217 & 1.987 & 0.006 & 0.194 & 1501 \\
50 & 0.215 & 0.753 & 0.007 & 0.128 & 607 \\
68 & 0.447 & 1.829 & 0.028 & 0.218 & 2044 \\
107 & 0.227 & 0.825 & 0.014 & 0.142 & 593 \\
124 & 0.285 & 0.966 & 0.009 & 0.170 & 865 \\
134 & 0.349 & 1.332 & 0.029 & 0.215 & 535 \\
151 & 0.299 & 1.022 & 0.010 & 0.196 & 898 \\
176 & 0.341 & 2.330 & 0.020 & 0.264 & 678 \\
185 & 0.281 & 0.922 & 0.021 & 0.187 & 799 \\
189 & 0.572 & 1.801 & 0.021 & 0.333 & 1467 \\
192 & 0.238 & 0.900 & 0.005 & 0.169 & 678 \\
194 & 0.284 & 1.450 & 0.011 & 0.215 & 747 \\
215 & 0.407 & 1.889 & 0.005 & 0.313 & 1056 \\
218 & 0.607 & 2.688 & 0.014 & 0.382 & 1223 \\
224 & 0.232 & 0.965 & 0.022 & 0.122 & 748 \\
230 & 0.403 & 1.423 & 0.005 & 0.277 & 866 \\
262 & 0.624 & 1.425 & 0.038 & 0.295 & 1365 \\
307 & 0.239 & 0.798 & 0.013 & 0.142 & 547 \\
308 & 0.178 & 0.651 & 0.009 & 0.117 & 516 \\
317 & 0.200 & 0.737 & 0.022 & 0.115 & 420 \\
318 & 0.363 & 1.409 & 0.016 & 0.223 & 551 \\
336 & 0.201 & 0.994 & 0.004 & 0.119 & 2759 \\
337 & 0.174 & 0.742 & 0.017 & 0.104 & 526 \\
344 & 0.118 & 0.504 & 0.009 & 0.083 & 530 \\
366 & 0.262 & 0.714 & 0.005 & 0.147 & 1671 \\
371 & 0.195 & 0.735 & 0.007 & 0.128 & 631 \\
376 & 0.244 & 0.919 & 0.011 & 0.148 & 808 \\
384 & 0.216 & 1.751 & 0.006 & 0.186 & 1313 \\
465 & 0.113 & 0.439 & 0.010 & 0.050 & 488 \\
469 & 0.072 & 0.300 & 0.001 & 0.051 & 903 \\
473 & 0.097 & 0.240 & 0.006 & 0.049 & 401 \\
476 & 0.100 & 0.424 & 0.003 & 0.052 & 922 \\
479 & 0.057 & 0.223 & 0.002 & 0.033 & 901 \\
487 & 0.254 & 0.956 & 0.005 & 0.144 & 722 \\
496 & 0.115 & 0.306 & 0.001 & 0.061 & 631 \\
498 & 0.105 & 0.277 & 0.009 & 0.050 & 494 \\
\bottomrule
\end{tabular}
\end{table}

\end{document}